\pdfoutput=1
\documentclass[11pt]{article}

\usepackage{acl}

\usepackage{times}
\usepackage{latexsym}

\usepackage[T1]{fontenc}
\usepackage[utf8]{inputenc}

\usepackage{microtype}

\usepackage{algorithm}
\usepackage{algorithmic}
\usepackage{amsmath}
\usepackage{multirow}
\usepackage{booktabs}
\usepackage{url}
\usepackage{graphicx}
\usepackage{enumitem}
\usepackage{amssymb}
\usepackage{bbm}
\usepackage{bm}
\usepackage{titlesec}
\usepackage{mathtools}
\usepackage{float}
\usepackage{arydshln}
\title{Query and Extract: Refining Event Extraction \\ as Type-oriented Binary Decoding}

\author{Sijia Wang$^{1}$,  Mo Yu$^{2}$,  Shiyu Chang$^{3}$, Lichao Sun$^{4}$, Lifu Huang$^{1}$
\\
  $^{1}$Virginia Tech 
  $^{2}$WeChat AI
  $^{3}$University of California Santa Barbara
  $^{4}$Lehigh University
 \\
  $^{1}${\tt \{sijiawang,lifuh\}@vt.edu}, 
  $^{2}${\tt moyumyu@tencent.com} \\
  $^{3}${\tt chang87@ucsb.edu},
  $^{4}${\tt lis221@lehigh.edu}
  }

\begin{document}
\maketitle

\begin{abstract}

Event extraction is typically modeled as a multi-class classification problem where event types and argument roles are treated as atomic symbols. These approaches are usually limited to a set of pre-defined types. We propose a novel event extraction framework that uses event types and argument roles as natural language queries to extract candidate triggers and arguments from the input text. With the rich semantics in the queries, our framework benefits from the attention mechanisms to better capture the semantic correlation between the event types or argument roles and the input text. Furthermore, the query-and-extract formulation allows our approach to leverage all available event annotations from various ontologies as a unified model. Experiments on ACE and ERE demonstrate that our approach achieves state-of-the-art performance on each dataset and significantly outperforms existing methods on zero-shot event extraction.\footnote{Our code is open
sourced at \url{https://github.com/VT-NLP/Event_Query_Extract} for reproduction purpose.}
\end{abstract}
\section{Introduction}
\label{sec:intro}

\begin{figure}
% \vspace{8mm}
  \centering
  \includegraphics[width=0.45\textwidth]{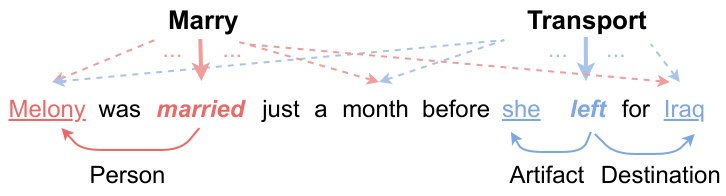}
  \caption{An example of event annotation.}
  \label{fig:intro_example}
\end{figure}

Event extraction~\cite{grishman1997information,chinchor1998muc,ahn2006stages} is a task to identify and type event triggers and participants from natural language text. As shown in
Figure~\ref{fig:intro_example}, \textit{married} and \textit{left} are triggers of two event mentions of the \textit{Marry} and \textit{Transport} event types respectively. Two arguments are involved in the \textit{left} event mention: \textit{she} is an \textit{Artifact}, and \textit{Irap} is the \textit{Destination}.

Traditional studies usually model event extraction as a multi-class classification 
problem~\cite{mcclosky2011event,qiliACl2013,chen2015event,yang-mitchell-2016-joint,nguyen_jrnn_2016,yinglinACL2020}, where a set of event types are first defined, and then supervised machine learning approaches will detect and classify each candidate event mention or argument into one of the target types. However, each event type or argument role is treated as an atomic symbol, ignoring their rich semantics in these approaches. Several studies explore the semantics of event types by leveraging the event type structures~\cite{huang2017zero}, seed event mentions~\cite{bronstein2015seed,lai2019extending}, or question answering (QA)~\cite{xinyaduEMNLP2020,jianliu2020emnlp}.
However, these approaches are still designed for and thus limited to a single target event ontology\footnote{An ontology is defined as a collection of event types and argument roles for a particular domain \cite{brown-etal-2017-rich, song2015light}.}, such as ACE~\cite{ldc_ace05} or ERE~\cite{song2015light}.

With the existence of multiple ontologies and the challenge of handling new emerging event types, it is necessary to study event extraction approaches that are generalizable and can use all available training data from distinct event ontologies.\footnote{For argument extraction, the QA-based approaches have certain potential to generalize to new ontologies, but require high-quality template questions. As shown in our experiments, their generalizability is limited compared to ours.}

\begin{figure*}
  \centering
  \includegraphics[width=0.91\textwidth]{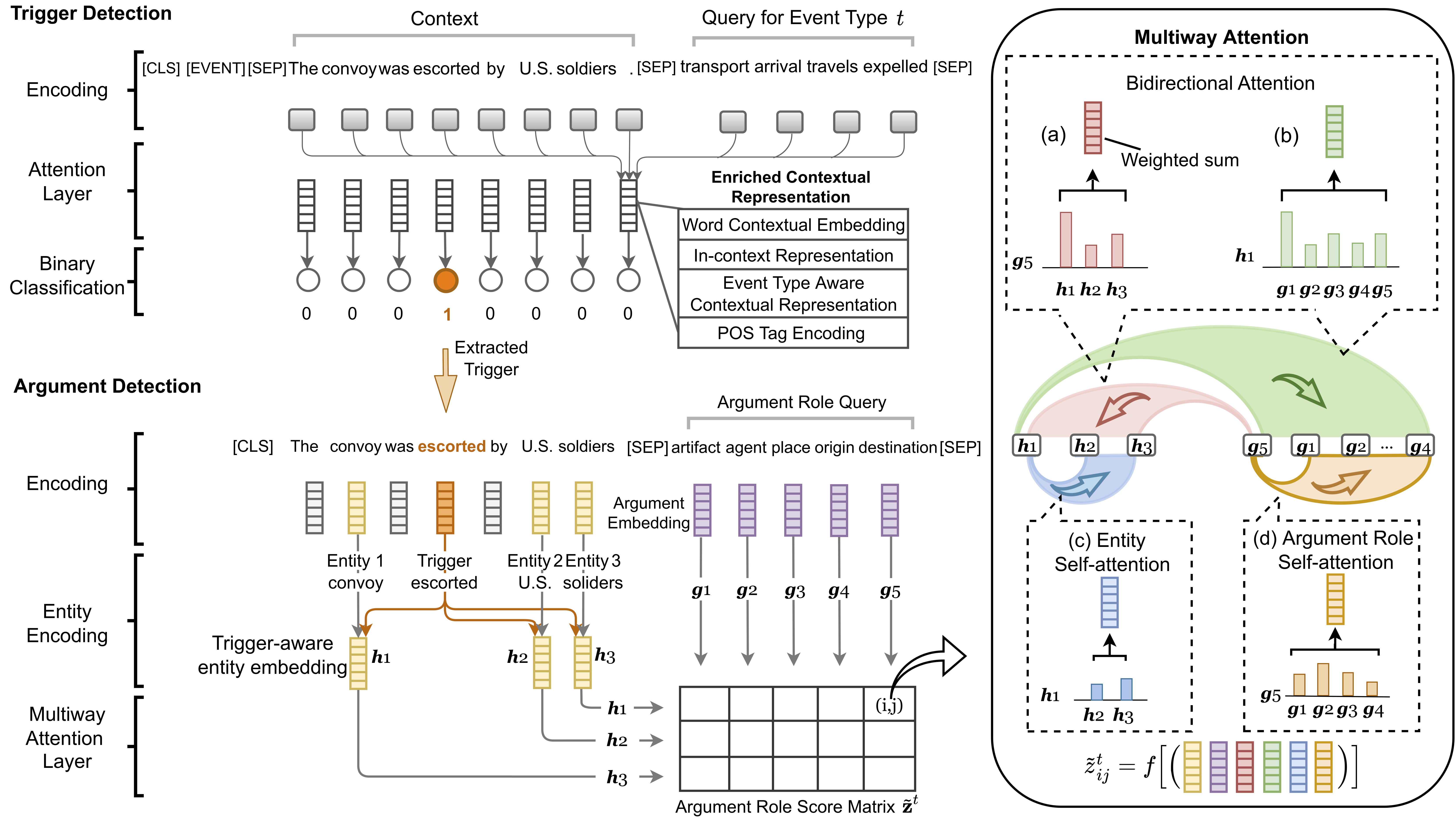}
  \caption{Architecture overview. Each cell in Argument Role Score Matrix indicates the probabilities of an entity being labeled with an argument role. The arrows in Multiway Attention module show four attention mechanisms: (a) entity to argument roles, (b) argument role to entities, (c) entity to entities, (d) argument role to argument roles.}
  \label{fig:architecture}
\end{figure*}

To this end,
we propose a new event extraction framework following a query-and-extract paradigm.
Our framework represents event types and argument roles as natural language queries with rich semantics.
The queries are then used to extract the corresponding event triggers and arguments by leveraging our proposed attention mechanism to capture their interactions with input texts. Specifically, (1) for trigger detection, we formulate each event type as a query based on its type name and a short list of prototype triggers, and make \textbf{binary decoding} of each token based on its query-aware embedding; (2) for argument extraction, we put together all argument roles defined under each event type as a query, followed by a multiway attention mechanism to extract all arguments of each event mention with \textbf{one-time encoding}, with each argument predicted as \textbf{binary decoding}.

Our approach can naturally handle various ontologies as a unified model --
compared to previous studies~\cite{nguyen-grishman-2016-modeling,wadden-etal-2019-entity,yinglinACL2020}, our binary decoding mechanism directly works with any event type or argument role represented as natural language queries, thus effectively leveraging cross-ontology event annotations and making zero-shot predictions.
Moreover, compared with the QA-based methods~\cite{xinyaduEMNLP2020,jianliu2020emnlp,EEMQA_li} that can also conduct zero-shot argument extraction, our approach does not require creating high-quality questions for argument roles or multi-time encoding for different argument roles separately, thus being more accurate and efficient.

We evaluate our approach on two public benchmark datasets, ACE and ERE, and demonstrate state-of-the-art performance in the standard supervised event extraction and the challenging transfer learning settings that generalize to new event types and ontologies. Notablely, on zero-shot transfer to new event types, our approach outperforms a strong baseline by 16\% on trigger detection and 26\% on argument detection. The overall contributions of our work are:

\begin{itemize}%[noitemsep,nolistsep,wide,itemindent=3pt]
 
\item We refine event extraction as a query-and-extract paradigm, which is more generalizable and efficient than previous top-down classification or QA-based approaches. 

\item We design a new event extraction model that leverages rich semantics of event types and argument roles, improving accuracy and generalizability. 

\item We establish new state-of-the-art performance on ACE and ERE in supervised and zero-shot event extraction and demonstrate our framework as an effective unified model for cross ontology transfer.
% and demonstrate our framework as a more efficient and accurate unified model that can make use of event annotations from different ontologies and extract events for any unseen types.
\end{itemize}

\section{Our Approach}
\label{sec:model}

%\lifu{add a figure to show the overall approach}
%\lifu{briefly describe the overview of the approach}

As Figure~\ref{fig:architecture} shows, given an input sentence, we first identify the candidate triggers for each event type by taking it as a query to the sentence. Each event type, such as \textit{Attack}, is represented with a natural language text, including its type name and a shortlist of prototype triggers, such as \textit{invaded} and \textit{airstrikes}, %\textit{overthrew}, and \textit{ambushed}, 
which are selected from the training examples. Then, we concatenate the input sentence with the event type query, encode them with a pre-trained BERT encoder~\cite{devlin-etal-2019-bert}, compute the attention distribution over the sequential representation of the event type query for each input token, and finally classify each token into a binary label, indicating it as a trigger candidate of the specific event type or not. 

To extract the arguments for each candidate trigger, we follow a similar strategy and take the set of pre-defined argument roles for its corresponding event type as a query to the input sentence. We use another BERT encoder to learn the contextual representations for the input sentence and the query of the argument roles. Then, we take each entity of the input sentence as a candidate argument and compute the semantic correlation between entities and argument roles with multiway attention, and finally classify each entity into a binary label in terms of each argument role.

\subsection{Trigger Detection}

%\lifu{use mathematical language to describe the process. Check previous papers, e.g., the OneIE paper}

% \lifu{you can divide the approach into several components, e.g., event type representation to describe how to get the anchor words and represent each type, context encoding, attention, event trigger prediction, etc. Add detailed formulas for each step}

\paragraph{Event Type Representation} A simple and intuitive way of representing an event type is to use the type name. However, the type name itself cannot accurately represent the semantics of the event type due to the ambiguity of the type name and the variety of the event mentions of each type. For example, \textit{Meet} can refer to \textit{an organized event} or an action of \textit{getting together} or \textit{matching}. Inspired by previous studies~\cite{bronstein2015seed,lai2019extending}, we use a short list of prototype triggers to enrich the semantics of each event type.
%, and more than 60 unique strings are annotated as \textit{Meet} event triggers in ACE05 dataset

%Previous studies either treat each event type as an atomic symbol~\cite{} or represent it with the type name only~\cite{jianliu2020emnlp, yinglinACL2020, xinyaduEMNLP2020}. Few studies explored seed triggers to denote the semantics of each event type and applied them to few-shot event extraction~\cite{}. Following similar intuitions, for each event type, we carefully select $K$ seed triggers from the training examples and use them to enrich the semantics of each event type.

%~\cite{jianliu2020emnlp, yinglinACL2020, xinyaduEMNLP2020} mainly rely on the type name to denote the semantics of each event type. 
%However, it's impossible for an event type name, which usually consists of two to three words, to cover the event type semantics. On the other hand, supplementary knowledge of each event type can be retrieved from the training data. For each event type, a collection of words or phrases are frequently labeled as triggers. Those words and phrases, named as seed triggers, are representative of the event type. 

%Thus we further select $K$ seed triggers $tr_1^{seed}\; tr_2^{seed}\; \dots tr_K^{seed} $  from the training examples based on two factors, term frequency and positive rate (in being tagged as an event trigger).

Specifically, for each event type $t$, we collect a set of annotated triggers from the training examples. For each unique trigger word, we compute its frequency from the whole training dataset as $f_o$ and its frequency of being tagged as an event trigger of type $t$ as $f_t$, and then obtain a probability $f_t/f_o$, which will be used to sort all the annotated triggers for event type $t$. We select the top-$K$\footnote{In our experiments, we set $K=4$.} ranked words as prototype triggers $\{\tau_1, \tau_2, \dots, \tau_K\}$. 

Finally, each event type will be represented with a natural language sequence of words, consisting of its type name and the list of prototype triggers $T= \{t, \tau_1^t, \tau_2^t, \dots, \tau_K^t\}$. Taking the event type \textit{Attack} as an example, we finally represent it as \textit{Attack invaded airstrikes overthrew ambushed}.

%each trigger mention, and its positive rate in being tagged as event trigger. We filter out less frequent trigger mentions, and select up to $K$ seed triggers with highest positive rate. \lifu{provide details: which numbers you computed, and how to rank them...}\sijia{ added}

%we sort the event trigger mentions from the training data by positive rates, and take the most frequent ones as seed triggers. For example, we choose \textit{invaded}, \textit{airstrikes}, \textit{overthrew}, and \textit{ambushed} as seed triggers for event type \textsc{attack}.

% \paragraph{Event Question Strategy.}
\paragraph{Context Encoding}
% \lifu{are you using POS tags? confirm}\sijia{I concatenate them before the last MLP layer}
Given an input sentence $W = \{w_{1}, w_{2}, \dots, w_{N} \}$, we take each event type $T= \{t, \tau_1^t, \tau_2^t, \dots, \tau_K^t\}$ as a query to extract the corresponding event triggers. Specifically, we first concatenate them into a sequence as follows:
\begin{gather*}
\text{[CLS]} \text{[EVENT]} \text{[SEP]}\; w_{1}\; ...\; w_{N}\; \text{[SEP]}\; \\t\; \tau_1^t\; ...\; \tau_K^t\;  \text{[SEP]}
\end{gather*}
% For each event type, we ask whether an token indicate such event by a question with the following three parts: 
where [SEP] is a separator from the BERT encoder~\cite{devlin-etal-2019-bert}. Following ~\cite{jianliu2020emnlp}, we use a special symbol [EVENT] to emphasis the trigger detection task. %as part of event type queries

Then we use a pre-trained BERT encoder to encode the whole sequence and get contextual representations for the input sentence $\boldsymbol{W} = \{\boldsymbol{w}_0, \boldsymbol{w}_2, ..., \boldsymbol{w}_N\}$ as well as the event type $ \boldsymbol{T} = \{\boldsymbol{t}, \boldsymbol{\tau}_0^t, \boldsymbol{\tau}_1^t, ..., \boldsymbol{\tau}_K^t\}$.\footnote{We use bold symbols to denote vectors.}

%The sequence connects the following three components  with special token [CLS] and [SEP] \cite{devlin-etal-2019-bert}:
%(i) a special token $\text{[EVENT]}$ to emphasis the trigger detection task. (ii) the context sequence $w_{1}\; w_{2}\; \dots w_{N}$. (iii) an event type query with event type name $ev_1 \;  ev_2$ and seed triggers $tr_1^{seed}\; tr_2^{seed}\; \dots tr_K^{seed} $. 
%For instance, as the example in Figure \ref{fig:architecture}, in the input sequence 
%"{[CLS] [EVENT] [SEP] The convoy was escorted by U.S. soldiers [SEP] transport arrival travels expelled [SEP]}", "transport" is the event sub-type token, and  "arrival travels expelled" are seed triggers. 
% \begin{displayquote}
% {[CLS] [EVENT] [SEP] Sentence [SEP] Event\_type Anchor\_words [SEP]}.
% \end{displayquote}
% where \textsc{[sep]} is the separator connecting the four parts, and \textsc{[cls]} is the special token indicate the start of the context. 

% \paragraph{Enriched Trigger Attention}
\paragraph{Enriched Contextual Representation}
%Different from previous studies~\cite{}, we aim to take each event type as query to an input sentence and detect candidate triggers by capturing the correlation between each token and the event type. 
Given a query of each event type, we aim to automatically extract corresponding event triggers from the input sentence. To achieve this goal, we need to capture the semantic correlation of each input token to the event type. Thus we apply attention mechanism to learn a weight distribution over the sequence of contextual representations of the event type query and get an event type aware contextual representation for each token:
%compute the relevance of each input token to the event type and learn an event type oriented contextual representation for each token.
%After obtaining the contextual representations, we aim to detect event triggers for each event type. To do so, we first learn an event type oriented contextual representation for each token based on the following attention mechanism:
% \begin{small}
\begin{gather*}
    \boldsymbol{A}_{i}^{T} = \frac{1}{T}\sum_{j=1}^{|T|}\alpha_{ij}\cdot\boldsymbol{T}_j \;,\\\alpha_{ij} = \cos(\boldsymbol{w}_i,\ \boldsymbol{T}_j)\;,
\end{gather*}
% \end{small}
where $\boldsymbol{T}_j$ is the contextual representation of the $j$-th token in the sequence $T= \{t, \tau_1^t, \tau_2^t, \dots, \tau_K^t\}$. $\cos(\cdot)$ is the cosine similarity function between two vectors. $\boldsymbol{A}_{i}^{T}$ denotes the event type $t$ aware contextual representation of token $w_i$.

In addition, the prediction of event triggers also depends on the occurrence of a particular context. For example, according to ACE event annotation guidelines~\cite{ldc_ace05}, to qualify as a \textit{Meet} event, the meeting must be known to be ``\textit{face-to-face and physically located somewhere}''. To capture such context information, we further apply in-context attention to capture the meaningful contextual words for each input token: 
%Taking \textit{Meet} event type as an example, some event types also require the occurrence of certain entities or context, for example, according to ACE2005 annotation guideline, to qualify as a \textit{Meet} Event, the meeting must be known to be face-to-face and physically located somewhere. To capture such context information, we further employ another in-context attention to capture meaningful contextual words for event type prediction: 
\begin{gather*}
    \boldsymbol{A}_{i}^W = \frac{1}{N}\sum_{j=1}^{N} \tilde{\alpha}_{ij}\cdot\boldsymbol{w}_j \; ,\\ \tilde{\alpha}_{ij} = \rho(\boldsymbol{w}_i,\ \boldsymbol{w}_j)\;,
\end{gather*}
where $\rho(.)$ is the attention function and is computed as the average of the self-attention weights from the last $m$ layers of BERT.\footnote{We set $m$ as 3 as it achieved the best performance.}

\paragraph{Event Trigger Detection}

With the event type oriented attention and in-context attention mechanisms, each token $w_i$ from the input sentence will obtain two enriched contextual representations $\boldsymbol{A}_{i}^W$ and $\boldsymbol{A}_{i}^{T}$. We concatenate them with the original contextual representation $\boldsymbol{w}_i$ from the BERT encoder, and classify it into a binary label, indicating it as a candidate trigger of event type $t$ or not:%\lifu{if you use pos tag features here, add it.}\sijia{added}
\begin{align*}
\small
   \boldsymbol{\tilde{y}}_i^t = \boldsymbol{U}_{o}\cdot([\boldsymbol{w}_i;\ \boldsymbol{A}_{i}^W;\ \boldsymbol{A}_{i}^{T}; \boldsymbol{P}_i]) \;,
\end{align*}
where $[;]$ denotes concatenation operation, $\boldsymbol{U}_{o}$ is a learnable parameter matrix for event trigger detection, and $\boldsymbol{P}_i$ is the one-hot part-of-speech (POS) encoding of word $w_i$. We optimize the following objective for event trigger detection
\begin{align*}
\small
    % L & = -\frac{1}{|\mathcal{T}|}\sum_{t\in\mathcal{T}}L^t_{trigge}\\
    \mathcal{L}_{1} = -\frac{1}{|\mathcal{T}||\mathcal{N}|}
        \sum_{t\in\mathcal{T}}
        \sum_{i=1}^{|\mathcal{N}|} \boldsymbol{y}_i^t\cdot\log \boldsymbol{\tilde{y}}_i^t\;,
\end{align*}
where $\mathcal{T}$ is the set of target event types and $\mathcal{N}$ is the set of tokens from the training dataset. $\boldsymbol{y}_i^t$ denotes the groundtruth label vector.

\subsection{Event Argument Extraction}
After detecting event triggers for each event type, we further extract their arguments based on the pre-defined argument roles of each event type.

\paragraph{Context Encoding}
Given a candidate trigger $r$ from the sentence $W = \{w_{1}, w_{2}, \dots, w_{N} \}$ and its event type $t$, we first obtain the set of pre-defined argument roles for event type $t$ as $G^t = \{g^t_1, g^t_2, ..., g^t_D\}$. To extract the corresponding arguments for $r$, similar as event trigger detection, we take all argument roles $G^t$ as a query and concatenate them with the original input sentence
\begin{align*}
    \text{[CLS]}\; w_{1}\; w_{2}\; ...\; w_{N}\; \text{[SEP]}\; g^t_1\; g^t_2\; ...\; g^t_D\; \text{[SEP]}
\end{align*}
where we use the last \text{[SEP]} separator to denote \textit{Other} category, indicating the entity is not an argument. 
Then, we encode the whole sequence with another pre-trained BERT encoder~\cite{devlin-etal-2019-bert} to get the contextual representations of the sentence $\boldsymbol{\tilde{W}} = \{\boldsymbol{\tilde{w}}_0, \boldsymbol{\tilde{w}}_2, ..., \boldsymbol{\tilde{w}}_N\}$, and the argument roles $ \boldsymbol{G}^t = \{\boldsymbol{g}_0^t, \boldsymbol{g}_1^t, ..., \boldsymbol{g}_D^t, \boldsymbol{g}_{\text{[Other]}}^t\}$.

As the candidate trigger $r$ may span multiple tokens within the sentence, we obtain its contextual representation $\boldsymbol{r}$ as the average of the contextual representations of all tokens within $r$. In addition, as the arguments are usually detected from the entities of sentence $W$, we apply a BERT-CRF model, which is optimized on the same training set as event extraction to identify the entities $E=\{e_1, e_2, ..., e_M\}$. As each entity may also span multiple tokens, following the same strategy, we average the contextual representations of all tokens within each entity and obtain the entity contextual representations as $\boldsymbol{E}=\{\boldsymbol{e}_1, \boldsymbol{e}_2, ..., \boldsymbol{e}_M\}$.

\paragraph{Multiway Attention} 

Given a candidate trigger $r$ of type $t$ and an entity $e_i$, for each argument role $g^t_{j}$, we need to determine whether the underlying relation between $r$ and $e_i$ corresponds to $g^t_{j}$ or not, namely, whether $e_i$ plays the argument role of $g^t_{j}$ in event mention $r$. To do this, for each $e_i$, we first obtain a trigger-aware entity representation as 
\begin{align*}
    \boldsymbol{h}_{i} = \boldsymbol{U}_{h}\cdot([\boldsymbol{e}_{i};\ \boldsymbol{r};\ \boldsymbol{e}_{i}\circ\boldsymbol{r}]) \;,
\end{align*}
where $\circ$ denotes element-wise multiplication operation. $\boldsymbol{U}_{h}$ is a learnable parameter matrix.

In order to determine the semantic correlation between each argument role and each entity, we first compute a similarity matrix $\boldsymbol{S}$ between the trigger-aware entity representations $\{\boldsymbol{h}_1, \boldsymbol{h}_2, ..., \boldsymbol{h}_M\}$ and the argument role representations $\{\boldsymbol{g}_0^t, \boldsymbol{g}_1^t, ..., \boldsymbol{g}_D^t\}$
\begin{align*}
    S_{ij} = \frac{1}{\sqrt{d}}\sigma(\boldsymbol{h}_{i},\   \boldsymbol{g}^{t}_{j})\;,
\end{align*}
where $\sigma$ denotes dot product operator, $d$ denotes embedding dimension of $\boldsymbol{g}^t$, and $S_{ij}$ indicates the semantic correlation of entity $e_i$ to a particular argument role $g^t_{j}$ given the candidate trigger $r$.

Based on the correlation matrix $\boldsymbol{S}$, we further apply a bidirectional attention mechanism to get an argument role aware contextual representation for each entity and an entity-aware contextual representation for each argument role as follows:  
\begin{gather*}
    \boldsymbol{A}^{e2g}_{i} = \sum_{j=1}^{D} \boldsymbol{S}_{ij}\cdot\boldsymbol{g}_{j}^{t}\;, \\
    \boldsymbol{A}^{g2e}_{j} = \sum_{i=1}^{M} \boldsymbol{S}_{ij}\cdot\boldsymbol{h}_{i}\;.
\end{gather*}

In addition, previous studies~\cite{hong2011using,qiliACl2013,yinglinACL2020} have revealed that the underlying relations among entities or argument roles are also important to extract the arguments. For example, if entity $e_1$ is predicted as \textit{Attacker} of an \textit{Attack} event and $e_1$ is \textit{located in} another entity $e_2$, it's very likely that $e_2$ plays an argument role of \textit{Place} for the \textit{Attack} event. To capture the underlying relations among the entities, we further compute the self-attention among them
\begin{align*}
    \mu_{ij} =\frac{1}{\sqrt d}\sigma(&\boldsymbol{h}_{i},\ \boldsymbol{h}_{j})  \;, \,\,\,\,\,\, \boldsymbol{\tilde{\mu}}_{i} = \text{Softmax}(\boldsymbol{\mu}_{i})\;, \\
    % & \boldsymbol{\tilde{\mu}}_{i} = \text{Softmax}(\boldsymbol{\mu}_{i})\;, \\
    &\boldsymbol{A}^{e2e}_{i} = \sum_{j=1}^{M}
    \tilde{\mu}_{ij}\cdot\boldsymbol{h}_{j}\;. 
    %  \boldsymbol{A}^{e2e}_{k} = &\frac{1}{|I_{e_k}|} \sum\nolimits_{j\in I_{e_k}}\sum\nolimits_i
    % {\mu}_{ij}\cdot\boldsymbol{h}_{i}\;,
    % \\
    % & \boldsymbol{A}^{e2e}_{i} = \boldsymbol{h}_{j}^{\top}\cdot\text{Softmax}(\boldsymbol{\mu}_{i})\;,
\end{align*}

Similarly, to capture the underlying relations among argument roles, we also compute the self-attention among them
\begin{align*}
    v_{jk} = \frac{1}{\sqrt{d}}\sigma(\boldsymbol{g}^{t}_{j} & ,\  \boldsymbol{g}^{t}_{k})\;,  \,\,\,\,\,\,
    \boldsymbol{\tilde{v}}_{j} = \text{Softmax}(\boldsymbol{v}_{j})\;, \\
    & \boldsymbol{A}^{g2g}_{j} = \sum_{k=1}^{D}
    \tilde{v}_{jk}\cdot\boldsymbol{g}_{k}^{t}\;.
\end{align*}

\paragraph{Event Argument Predication} Finally, for each candidate event trigger $r$, we determine whether an entity $e_i$ plays an argument role of $g^{t}_{j}$ in the event mention by classifying it into a binary class:
\begin{align*}
    \boldsymbol{\tilde{z}}^{t}_{ij} = \boldsymbol{U}_{a}\cdot([\boldsymbol{h}_{i};\ \boldsymbol{g}^{t}_{j};\ \boldsymbol{A}^{e2g}_{i};\ \boldsymbol{A}^{g2e}_{j};\ \boldsymbol{A}^{e2e}_{i};\ \boldsymbol{A}^{g2g}_{j}]) ,
\end{align*}
where $\boldsymbol{U}_{a}$ is a learnable parameter matrix for argument extraction. And $\tilde z^t$ is argument role score matrix for event type $t$. The training objective is to minimize the following loss function:
\begin{align*}
   \mathcal{L}_{2} = -\frac{1}{|\mathcal{A}||\mathcal{E}|}
        \sum_{j=1}^{|\mathcal{A}|}
        \sum_{i=1}^{|\mathcal{E}|} \boldsymbol{z}_{ij} \log \boldsymbol{\tilde{z}}_{ij}\;,
\end{align*} 
where $\mathcal{A}$ denotes the collection of possible argument roles, and $\mathcal{E}$ is the set of entities we need to consider for argument extraction. $\boldsymbol{z}_{ij}$ denotes the ground truth label vector. During test, an entity will be labeled as a non-argument if the prediction for \textit{Other} category is 1. Otherwise, it can be labeled with multiple argument roles.

\section{Experiments}
\label{sec:exp}

\begin{table*}
\centering
\small
\resizebox{0.8\textwidth}{!}
{
\begin{tabular}{l|c|c|c|c}
\toprule
\multirow{2}{*}{Model}&  
\multicolumn{2}{c|}{ACE05-E$^+$} &  
\multicolumn{2}{c}{ERE-EN}
\\
\cmidrule{2-5}
&Trigger Ext. & Argument Ext. & Trigger Ext. & Argument Ext. 
 \\
\midrule
    %JointBeam \cite{qiliACl2013} & 67.5 & 41.8 & 52.7 & - \\
    % JRNN \cite{nguyen_jrnn_2016} &69.3\\
    % dbRNN~\cite{dbrnn_sha_2018} & 69.6 & 50.1 & 58.7 & - \\
    DYGIE++~\cite{wadden-etal-2019-entity}  & 67.3$^*$ & 42.7$^*$ & - & - \\
    BERT\_QA\_Arg~\cite{xinyaduEMNLP2020}&70.6$^*$ & 48.3$^*$ & 57.0 & 39.2\\
    OneIE~\cite{yinglinACL2020} & 72.8 & 54.8 & 57.0  & 46.5 \\
    Text2Event~\cite{text2event} &71.8 & 54.4 & 59.4 & 48.3 \\
    FourIE~\cite{vannguyen2021crosstask}&73.3 & \textbf{57.5} &  57.9  & 48.6 \\
    % MQAEE ~\cite{EEMQA_li} &71.7 & 53.4 & - & - \\
    %RCEE \cite{jianliu2020emnlp} & 74.9*? & 58.7? & ? & ? \\
\midrule    
    % \textbf{Query\&Extract} & \textbf{73.7} & 55.1 & \textbf{59.7} & \textbf{67.2}\\
    % Our Approach w/o (Multiway) Attention&& & 72.2 & 54.0  & 55.6  & 42.2 \\
    % \textbf{Our Approach}& \textbf{73.7} & 55.1(55.6) &  \textbf{60.4}  & \textbf{50.2} (50.8)\\
        \textbf{Our Approach} & \textbf{73.6} (0.2) & 55.1 (0.5) &  \textbf{60.4} (0.3) & \textbf{50.4} (0.3)\\
\bottomrule
\end{tabular}
}
\caption{Event extraction results on ACE05-E$^+$ and ERE-EN datasets (F-score, \%). $^*$ indicates scores obtained from their released codes. The performance of BERT\_QA\_Arg is lower than that reported in \cite{xinyaduEMNLP2020} as they only consider single-token event triggers. Each score of our approach is the mean of three runs and the variance is shown in parenthesis.
% Argument Ext. (GE) indicates the argument extraction results with gold entities and system predicted triggers. Argument Ext. (GT) indicates the argument extraction performance with gold triggers and system detected entities. 
% *indicate with data augmentation. 
%$^{\star}$ indicates statistical significance $(p < 0.05)$.
%\sijia{$^*$ The result is reproduced by the released code}
}
\label{tab:ACE}
\end{table*}

\subsection{Experimental Setup}
%\paragraph{Dataset} 
We perform experiments on two public benchmarks, ACE05-E$^+$\footnote{\url{https://catalog.ldc.upenn.edu/LDC2006T06}} and ERE-EN~\cite{song2015light}\footnote{Following~\citet{yinglinACL2020}, we merge LDC2015E29, LDC2015E68, and LDC2015E78 as the ERE dataset.}.
%Automatic Content Extraction 2005 (ACE05-E$^+$)\footnote{\url{https://catalog.ldc.upenn.edu/LDC2006T06}} and Entity Relation Event (ERE-EN)~\cite{song2015light}\footnote{Following~\citet{yinglinACL2020}, we merge LDC2015E29, LDC2015E68, and LDC2015E78 as the ERE dataset.}. 
ACE defines 33 event types while ERE includes 38 types, among which there are 31 overlapped event types. We use the same data split of ACE and ERE as~\cite{wadden-etal-2019-entity,yinglinACL2020,xinyaduEMNLP2020} for supervised event extraction. For zero-shot event extraction, we use the top-$10$ most popular event types in ACE as seen types for training and treat the remaining 23 event types as unseen for testing, following~\citet{huang2017zero}. In our experiments, we use random seeds
and report averaged scores of each setting. More details regarding the data statistics and evaluation are shown in Appendix~\ref{app:data}.

We further design two more challenging and practical settings to evaluate how well the approach could leverage resources from different ontologies: (1) \textit{cross-ontology direct transfer}, where we only use the annotations from ACE or ERE for training and directly test the model on another event ontology. This corresponds to the \emph{domain adaptation setting} in transfer learning literature; %This is a more challenging task than the ACE05 zero-shot transfer task described earlier;
(2) \textit{joint-ontology enhancement}, where we take the annotations from both ACE and ERE as the training set and test the approaches on ACE or ERE ontology separately. This corresponds to the \emph{multi-domain learning setting} in transfer learning literature.
Intuitively, an approach with good transferability should benefit more from the enhanced training data from other ontologies. We follow the same train/dev/test splits of ACE and ERE as supervised event extraction.

\subsection{Supervised Event Extraction}
\label{sec:sup_event_ext}
%Our first experiment is the standard supervised event extraction on both ACE and ERE.
%Under supervised learning scenario, we leverage annotated data of all the event types.

Table~\ref{tab:ACE} shows the supervised event extraction results of various approaches on ACE and ERE datasets. Though studies~\cite{yang-mitchell-2016-joint,jianliu2020emnlp,xiaoliuACL2018, dbrnn_sha_2018, lai2020emnlp, Veyseh2020EMNLP} have been conducted on the ACE dataset, they follow different settings\footnote{Many studies did not describe their argument extraction setting in detail.}, especially regarding whether the Time and Value arguments are considered and whether all Time-related argument roles are viewed as a single role. Following several recent state-of-the-art studies~\cite{wadden-etal-2019-entity,yinglinACL2020,xinyaduEMNLP2020}, we do not consider Time and Value arguments. Our approach significantly outperforms most of the previous comparable baseline methods, especially on the ERE dataset\footnote{Appendix~\ref{sec:remaining_challenges} describes several remaining challenges identified from the prediction errors on ACE05 dataset.}. Next, we take BERT\_QA\_Arg, a QA\_based method, as the main baseline as it shares similar ideas to our approach to compare their performance. 

%on trigger detection and argument extraction
%The comparison between our approach and the ablated variant without attention mechanisms demonstrates the effectiveness of the attention mechanisms in capturing the correlation between input tokens and event type or argument role queries. 
%To better illustrate the advantage of our approach, we also provide the argument extraction performance based on gold entities or gold triggers as shown in Table~\ref{tab:ACE}. 
%\sijia{Without attention mechanisms, the model will be reduced to the adapted BERT\_QA\_Arg baseline, which shows the impact of our attention mechanisms.} 

% \lifu{emphasize why we don't compare with other baselines}
% \sijia{Previous studies differ in argument detection setting from two aspects, whether the Time and Value arguments are included and whether all Time-related argument roles are viewed as a single role. We follow several state-of-the-art studies and ignore Value and Time arguments. This setting is different from the studies (including ~\cite{jianliu2020emnlp} and ~\cite{xiaoliuACL2018}). \citet{lai2020emnlp} only focuses on trigger detection, while \cite{Veyseh2020EMNLP} only focuses on argument extraction and leveraged external NYT corpus \cite{wang-etal-2019-adversarial-training}. }

%\sijia{Cite , \cite{feng-etal-2016-language}, (?pre for event class > pre for event identification) \cite{cao-etal-2015-improving}, \cite{cleve}}

%\lifu{add the results on MAVEN event detection}\cite{MAVEN}

Specifically, for trigger detection, all the baseline methods treat the event types as symbols and classify each input token into one of the target types or \textit{Other}. So they heavily rely on human annotations and do not perform well when the annotations are not enough. For example, there are only 31 annotated event mentions for \textit{End\_Org} in the ACE05 training dataset, so BERT\_QA\_Arg only achieves 35.3\% F-score. In comparison, our approach leverages the semantic interaction between the input tokens and the event types. Therefore it still performs well when the annotations are limited, e.g., it achieves 66.7\% F-score for \textit{End\_Org}. In addition, by leveraging the rich semantics of event types, our approach also successfully detects event triggers that are rarely seen in the training dataset, e.g., \textit{ousting} and \textit{purge} of \textit{End-Position}, while BERT\_QA\_Arg misses all these triggers. A more detailed discussion about the impact of seed triggers is in Appendix~\ref{sec:impact_seed}.

For argument extraction, our approach shows more consistent results than baseline methods. For example, in the sentence ``\textit{Shalom was to fly on to London for talks with British Prime Minister Tony Blair and Foreign Secretary Jack Straw}'', the BERT\_QA\_Arg method correctly predicts \textit{Tony Blair} and \textit{Jack Straw} as \textit{Entity} arguments of the \textit{Meet} event triggered by \textit{talks}, but misses \textit{Shalom}. However, by employing multiway attention, especially the self-attention among all the entities, our approach can capture their underlying semantic relations, e.g., \textit{Shalom} and \textit{Tony Blair} are two persons to talk, so that it successfully predicts all the three \textit{Entity} arguments for the \textit{Meet} event. 

\begin{table}
\centering
\small
\resizebox{0.45\textwidth}{!}
{
\begin{tabular}{p{2.7cm}|c|c}
\toprule
Model & Trigger Ext. & Arg Ext. (GT) \\ 
\midrule
%Zero-Shot~\cite{huang2017zero} & 33.5 & 14.7 \\
%RCEE~\cite{jianliu2020emnlp} & 5.33 $\dagger$ & ? \\
% BERT\_QA\_Arg \cite{xinyaduEMNLP2020}$^\dagger$ & \multirow{2}{*}{31.6} & \multirow{2}{*}{17.0} \\
BERT\_QA\_Arg$^\dagger$ & 31.6 & 17.0 \\
\midrule
\textbf{Our Approach} & \textbf{47.8} &  \textbf{43.0}   \\
% \midrule
\bottomrule
\end{tabular}
}
\caption{Zero-shot F-scores on 23 unseen event types. $\dagger$: adapted implementation from \cite{xinyaduEMNLP2020}. GT indicates using gold-standard triggers as input.}
\label{tab:zeroshot}
\end{table}

\subsection{Zero-Shot Event Extraction}
\label{subsec:zero-shot}

\begin{table*}[t!]
\centering
\footnotesize
\resizebox{0.95\textwidth}{!}
{
\begin{tabular}{l|c|cc|cc|cc}
\toprule
\multirow{2}{*}{Source} & \multirow{2}{*}{Target} &  \multicolumn{2}{c|}{BERT\_QA\_Arg$_{\textrm{multi}}$}&
\multicolumn{2}{c|}{BERT\_QA\_Arg$_{\textrm{binary}}$$\dagger$} &\multicolumn{2}{c}{\textbf{Our Approach}}  \\ 
\cmidrule{3-8}
 &  & Trigger Ext. & Argument Ext. & Trigger Ext. & Argument Ext. & Trigger Ext. & Argument Ext.  \\ 
\midrule
ERE & ACE    & 48.9 (48.9)  &  18.5 (18.5)   & 50.8 (50.8) & 20.9 (20.9) & 53.9 (52.6)  & 30.2 (29.6) \\
ACE & ACE    & 70.6 & 48.3  & 72.2  & 50.4  & 73.6 &  55.1 \\
% MAVEN & ACE & & -&& -& 12.4 & -\\
% ACE+ERE & ACE & 70.1 & 47.0  & 71.3 & 49.8 & 74.4 (56.3) & 56.2 (41.9) \\ \midrule
ACE+ERE & ACE & 70.1 & 47.0  & 71.3 & 49.8 & 74.4 & 56.2 \\ \midrule
ACE & ERE    &47.2 (47.2) & 18.0 (18.0)  & 47.2 (45.0) & 17.9 (17.1) & 55.9 (46.3) & 31.9 (26.0) \\
ERE & ERE    & 57.0  & 39.2   & 56.7  &  42.9 & 60.4  &  50.4 \\
% ACE+ERE & ERE  &57.0  &  38.6 & 54.6 & 37.1 & 63.1 (51.5) & 52.3 (42.8) \\
ACE+ERE & ERE  &57.0  &  38.6 & 54.6 & 37.1 & 63.0 & 52.3 \\
% \midrule
% ACE & MAVEN     & & - & & - & & -\\
% MAVEN & MAVEN   & & - & & - & 68.8 & - \\
% ACE+MAVEN & MAVEN & & & & & & \\
\bottomrule
\end{tabular}
}
\caption{Cross ontology transfer between ACE and ERE datasets (F-score \%).
%In each block, the first row corresponds to the direct transfer setting and the third corresponds to the enhancement setting.
%We list the in-ontology results (second row) for reference.
The scores in parenthesis indicate the performance on the ACE and ERE shared event types. }
%\lifu{In parenthesis, also add the performance on the ACE/ERE shared types on test datasets} \lifu{Reviewers may also ask what happen if we only use annotations for the overlapped types}
\label{tab:crossontology}
\end{table*}

%As our approach makes great use of the semantics of both event types and argument roles, it's naturally capable of extracting event mentions and arguments for the types that are unseen during training, thus we demonstrate the transfer learning capability of our approach with zero-shot event extraction.

As there are no fully comparable baseline methods for zero-shot event extraction, we adapt one of the most recent states of the arts, BERT\_QA\_Arg~\cite{xinyaduEMNLP2020}, which is expected to have specific transferability due to its QA formulation.
However, the original BERT\_QA\_Arg utilizes a generic query, e.g., ``\textit{trigger}'' or ``\textit{verb}'', to classify each input token into one of the target event types or \textit{Other}, thus is not capable of detecting event mentions for any new event types during the test. We adapt the BERT\_QA\_Arg framework by taking each event type instead of the generic words as a query for event detection. Note that our approach utilizes the event types as queries without prototype triggers for zero-shot event extraction. 

As Table~\ref{tab:zeroshot} shows, our approach significantly outperforms BERT\_QA\_Arg under zero-shot event extraction, with over 16\% F-score gain on trigger detection and 26\% F-score gain on argument extraction. Comparing with BERT\_QA\_Arg, which only relies on the self-attention from the BERT encoder to learn the correlation between the input tokens and the event types or argument roles, our approach further applies multiple carefully designed attention mechanisms over BERT contextual representations to better capture the semantic interaction between event types or argument roles and input tokens, yielding much better accuracy and generalizability. 

% The main difference between our approach and BERT\_QA\_Arg lies in the attention mechanism used for trigger and argument extraction. For example, BERT\_QA\_Arg relies on the self-attention within BERT encoder to learn the correlation between event types or argument roles and each input token, while our approach further applies multiple carefully-designed attention mechanisms over BERT contextual representations to better capture the semantic interactions between event types or argument roles and each input token, and among the set of entities or argument roles.

\begin{figure}
  \centering
  \includegraphics[width=.45\textwidth]{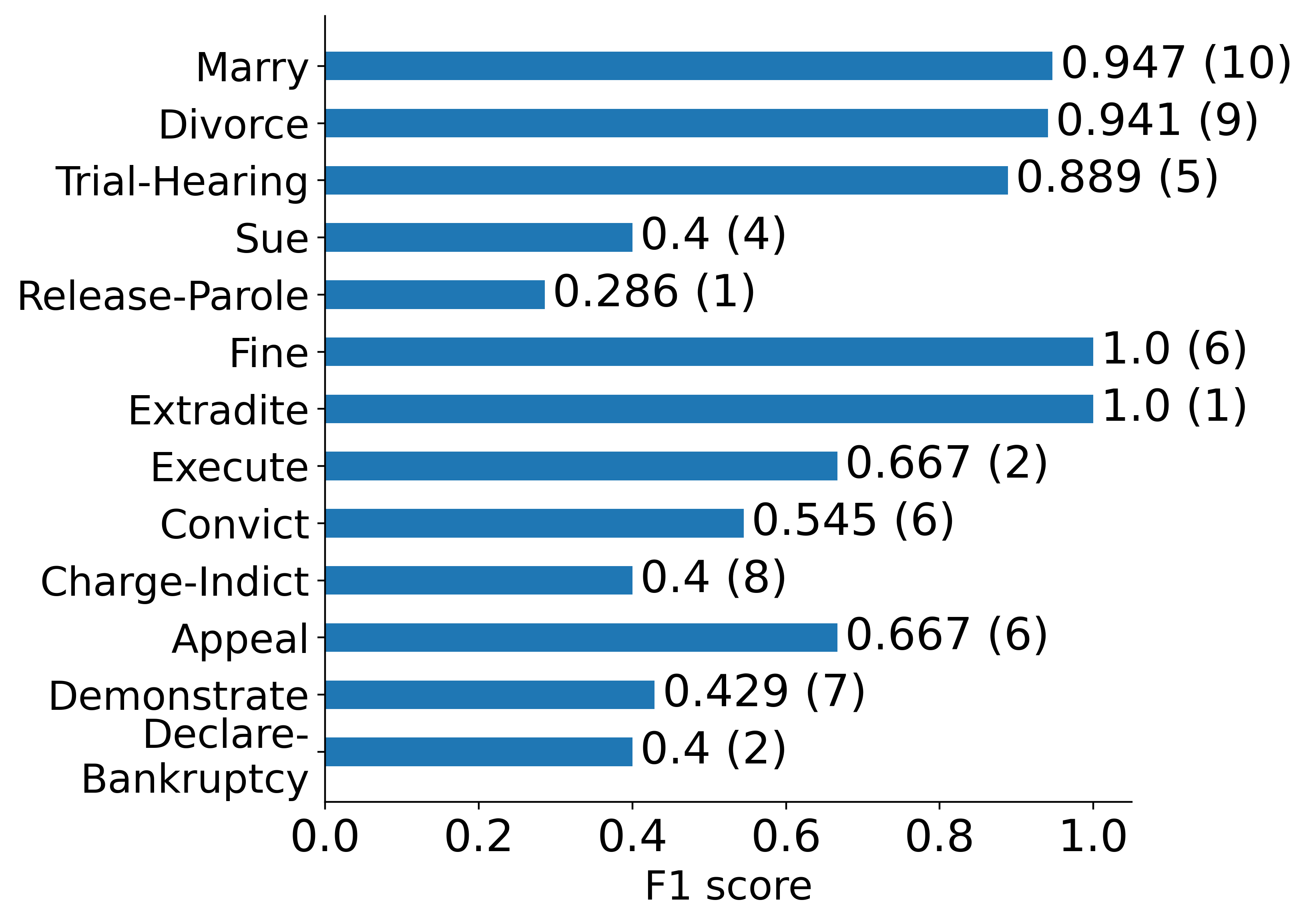}
  \caption{Zero-shot event extraction on each unseen event type. The number in parenthesis indicates \# gold event mentions of each unseen type in the test set.}%\sijia{redraw the figure, sort the f1 score in decreasing order}}
  \label{fig:zero_shot_error}
\end{figure}

We further pick 13 unseen event types and analyze our approach's zero-shot event extraction performance on each of them. As shown in Figure~\ref{fig:zero_shot_error}, our approach performs exceptionally well on \textit{Marry}, \textit{Divorce}, \textit{Trial-Hearing}, and \textit{Fine}, but worse on \textit{Sue}, \textit{Release-Parole}, \textit{Charge-Indict}, \textit{Demonstrate}, and \textit{Declare-Bankruptcy}, with two possible reasons: first, the semantics of event types, such as \textit{Marry}, \textit{Divorce}, is more straightforward and explicit than other types, such as \textit{Charge-Indict}, \textit{Declare-Bankruptcy}. Thus our approach can better interpret these types. Second, the diversity of the event triggers for some types, e.g., \textit{Divorce}, is much lower than other types, e.g., \textit{Demonstrate}. For example, among the 9 \textit{Divorce} event triggers, there are only 2 unique strings, i.e., \textit{divorce} and \textit{breakdowns}, while there are 6 unique strings among the 7 event mentions of \textit{Demonstrate}.

\subsection{Cross Ontology Transfer}

For cross-ontology transfer, we develop two variations of BERT\_QA\_Arg as baseline methods: (1) BERT\_QA\_Arg$_{\textrm{multi}}$, which is the same as the original implementation and use multi-classification to detect event triggers. (2) BERT\_QA\_Arg$_{\textrm{binary}}$, for which we apply the same query adaptation as Section~\ref{subsec:zero-shot} and use multiple binary-classification for event detection. For \textit{joint-ontology enhancement}, we combine the training datasets of ACE and ERE and optimize the models from scratch.\footnote{Another intuitive training strategy is to train the model on the source and target ontologies sequentially. Our pilot study shows that this strategy performs slightly worse.}

% we also explore two strategies: one is to directly train the models on the combination of the annotations from ACE and ERE, and the second strategy is to first pre-train a model on the ontology that is different from the target one, and then fine-tune the model on the annotations of the target ontology. We only report the performance based on the first ontology, which shows better performance in our experiments.\lifu{confirm}\sijia{do yo mean the performance of our approach} 

Table~\ref{tab:crossontology} shows the cross-ontology transfer results in both \emph{direct transfer} and \emph{enhancement} settings. Our approach significantly outperforms the baseline methods under all the settings. Notably, for \emph{direct transfer}, e.g., from ERE to ACE, by comparing the F-scores on the whole test set with the performance on the ACE and ERE shared event types (F-scores shown in parenthesis), our approach not only achieves better performance on the shared event types but also extracts event triggers and arguments for the new event types in ACE. In contrast, the baseline methods hardly extract any events or arguments for the new event types. 
Moreover, by combining the training datasets of ACE and ERE for \textit{joint-ontology enhancement}, our approach's performance can be further boosted compared with using the annotations of the target event ontology only, demonstrating the superior transfer capability across different ontologies. For example, ACE includes a \textit{Transport} event type while ERE defines two more fine-grained types \textit{Transport-Person} and \textit{Transport-Artifact}. By adding the annotations of \textit{Transport-Person} and \textit{Transport-Artifact} from ERE into ACE, our approach can capture the underlying semantic interaction between \textit{Transport}-related event type queries and the corresponding input tokens and thus yield 1.5\% F-score gain on the \textit{Transport} event type of ACE test set. In contrast, both baseline methods fail to be enhanced with additional annotations from a slightly different event ontology without explicitly capturing semantic interaction between event types and input tokens. Appendix~\ref{appendix:cross} provides a more in-depth comparison between our approach and the baseline approaches.
%The only difference between our approach and BERT\_QA\_Arg$_{\textrm{binary}}$ for trigger detection lies in the use of seed triggers in event queries and the
%\lifu{need to add reasons based on the results}

% \sijia{Comparing with MRC-based methods, our approach has two advantages: (1) we use a query that only contains event type-related information and let the model compare the semantics of the input token with the query. In contrast, for QA-based methods, the question consists of noisy information besides the event type like “which is” or “what is.” The model can better find the semantically aligned words with the query than with a question in natural language. That is why our approach performs better than QA-based methods. (2) To improve the semantically matching between the input tokens and the query, we designed several attention mechanisms to capture the contextual correlation of each input token and the query. (3) Many studies \cite{xinyaduEMNLP2020, jianliu2020emnlp} have shown that MRC models are sensitive to the subtle change of the questions, which explains the performance drop of MRC-based methods when tackling event annotations from a different ontology.}

\subsection{Ablation Study}
\label{sec:Ablation}
We further evaluate the impact of each attention mechanism on event trigger detection and argument extraction. As Table~\ref{tab:Ablation} shows, all the attention mechanisms show significant benefit to trigger or argument extraction, especially on the ERE dataset. The Event Type Attention and Multiway Attention show the most effects to trigger and argument extraction, which is understandable as they are designed to capture the correlation between the input texts and the event type or argument role-based queries. We also notice that, without taking entities detected by the BERT-CRF name tagging model as input, but instead considering all the tokens as candidate arguments\footnote{We take consecutive tokens predicted with the same argument role as a single argument span.}, our approach still shows competitive performance for argument extraction compared with the strong baselines. More ablation studies are discussed in Appendix~\ref{sec:more_ablation}.

\begin{table}[h]\centering
\footnotesize
\begin{tabular}{l|l|c|c}
\toprule
& Model & ACE & ERE \\ 
\midrule
\multirow{4}{*}{Trigger} & Our Approach & 73.6 & 60.4\\
& w/o Seed Trigger & 72.2 & 58.2 \\
& w/o In-Context Attention & 72.3 & 57.9\\
& w/o Event Type Attention & 71.1 & 56.9\\
\midrule
\multirow{5}{*}{Arg.} & Our Approach & 55.1 & 50.4\\
%w/o Trigger Info & 51.9 & \\
& w/o Entity Detection & 53.0 & 47.6 \\
& w/o Multiway Attention &  53.4  & 42.8 \\
& w/o Entity Self-attention & 53.7 & 48.3\\
& w/o Arg Role Self-attention & 54.1 & 47.7\\
\bottomrule
\end{tabular}
\caption{Results of various ablation studies. Each score is the average of three runs for each experiment.}
\label{tab:Ablation}
\end{table}

\subsection{Pros and Cons of Type-oriented Decoding}
The advantages of our type-oriented binary decoding include: (1) it allows the model to better leverage the semantics of event types which have been proved effective for both supervised and zero-shot event extraction; 
(2) it allows the approach to leverage all available event annotations from distinct ontologies, which is demonstrated in zero-shot event extraction and cross-ontology transfer; (3) in practice, new event types and annotations could emerge incessantly, and it is not possible to always train a model for all the event types. Our approach has the potential to be continuously updated and extract events for any desired event types. 

We also admit that binary decoding usually increases the computation cost. We design two strategies to mitigate this issue: (1) More than 69\% of the sentences in the training dataset do not contain any event triggers, so we randomly sample 20\% of them for training. 
(2) Our one-time argument encoding and decoding strategies extract all arguments of each event trigger at once. It is more efficient than the previous QA-based approaches, which only extract arguments for one argument role at once. With these strategies, for trigger detection, our approach takes 80\% more time for training and 19\% less for inference compared with BERT\_QA\_Arg which relies on multi-class classification, while for argument extraction, our approach takes 36\% less time for training and inference than BERT\_QA\_Arg. Even on a more fine-grained event ontology MAVEN~\cite{MAVEN}, which consists of 168 event types, for trigger extraction, our approach roughly takes 20\% more time for training and twice the time for inference compared with BERT\_QA\_Arg, with slightly better performance than the state of the art~\cite{cleve}\footnote{Our approach achieves 68.8\% F-score on MAVEN. We do not discuss more as MAVEN only contains trigger annotations.}.

\section{Related Work}
\label{sec:related}
% should be argument detection instead of argument classification 
% ACE has overlap arguments 
% ERE has overlap triggers and arguments 
% The evaluation is spurious
%
% In this section, we briefly describe prior work from the areas of event extraction and heterogeneous graph embedding that are most related to our work.

% Since single candidate span may act as triggers for multiple event occurrences, traditional frameworks that separates trigger identification and classification suffer from systematic error. 

%Event extraction has been a long-studied and challenging task in natural language processing (NLP)~\cite{Riloff1996, qiliACl2013, wadden-etal-2019-entity}. 

% , based on hand-crafted features or deep embedding based features~\cite{pennington2014glove,peters2018deep,devlin-etal-2019-bert}

Traditional event extraction studies~\cite{mcclosky2011event,qiliACl2013,chen2015event,cao-etal-2015-improving,feng-etal-2016-language,yang-mitchell-2016-joint,nguyen_jrnn_2016,zhang2017improving,wadden-etal-2019-entity,yinglinACL2020,cleve} usually detect event triggers and arguments with multi-class classifiers. Unlike all these methods that treat event types and argument roles as symbols, our approach considers them queries with rich semantics and leverages the semantic interaction between input tokens and each event type or argument role.

Several studies have explored the semantics of event types based on seed event triggers~\cite{bronstein2015seed,lai2019extending,zhang-etal-2021-zero}, event type structures~\cite{huang2016liberal,huang2017zero}, definitions~\cite{chen2019reading} and latent representations~\cite{huang2020semi}. However, they can hardly be generalized to argument extraction. Question answering based event extraction~\cite{xinyaduEMNLP2020, jianliu2020emnlp, EEMQA_li, Lyu-etal-2021-zero} can take advantage of the semantics of event types and the large-scale question answering datasets. Compared with these methods, there are three different vital designs, making our approach perform and be generalized better than these QA-based approaches: (1) our approach directly takes event types and argument roles as queries. In contrast, previous QA-based approaches rely on templates or generative modules to create natural language questions. However, it is difficult to find the optimal format of questions for each event type, and many studies \cite{xinyaduEMNLP2020, li-etal-2020-unified} have shown that MRC or QA models are sensitive to the subtle change of the questions.
(2) QA-based approaches can only detect arguments for one argument role at once, while our approach extracts all arguments of an event trigger with one-time encoding and decoding, which is more efficient and leverages the implicit relations among the candidate arguments or argument roles. (3) QA-based approaches rely on span prediction to extract arguments without requiring entity extraction, which could result in more entity boundary errors. Thus we choose to pre-train a name tagging model and use binary decoding over system detected entities.Moreover, it is pretty challenging to fully adapt the event extraction task to the span-based Question Answering. The main reason is that each sentence may contain multiple triggers for a particular event type. Even if we can formalize a question, e.g., “what is the trigger for Attack?” it is difficult for the model to return all the spans of event triggers correctly.

\section{Conclusion and Future Work}
\label{sec:conclusion}
% In this paper, we presented a novel graph neural network based framework - $\emph{SemPathNet}$ - an end-to-end modeling framework for heterogeneous graph embedding. Our experiments are conducted on two scenarios, i.e., node classification and node clustering. The experimental results illustrate that $\emph{SemPathNet}$ outperforms various state-of-the-art models.

We refine event extraction with a query-and-extract paradigm and design a new framework that leverages rich semantics of event types and argument roles and captures their interactions with input texts using attention mechanisms to extract event triggers and arguments. Experimental results demonstrate that our approach achieves state-of-the-art performance on supervised event extraction and shows prominent accuracy and generalizability to new event types and across ontologies. In the future, we will explore better representations of event types and argument roles and combine them prompt tuning approach further to improve the accuracy and generalizability of event extraction.

%\sijia{ We plan to summarize previous different settings for a fair comparison.
%From various efforts on Event Extraction tasks, we found that a standard evaluation method is missing. Previous work varies in dataset pre-processing and task formulations, such as the argument types considered. We also plan to investigate the effect of different query strategies. The proposed event representation is representative but sub-optimal. Seed triggers are semantically meaningful and can align nicely with high-frequency trigger candidates. However, they could boost the possibility of making a false positive prediction on related tokens and omitting unseen or rare candidates.}

%extend our framework to other information extraction tasks and to leverage available annotations from related tasks.

%Experiments on supervised event extraction, zero-shot event extraction as well as cross-ontology transfer with two public benchmark datasets ACE and ERE demonstrate that our approach not only yields more accurate 

\section*{Acknowledgements}
We thank the anonymous reviewers and area chair for their valuable time and constructive comments, and the helpful discussions with Zhiyang Xu and Minqian Liu. We also thank the support from the Amazon Research Awards.

\bibliography{sijia_ref}
\bibliographystyle{acl_natbib}

\appendix
\section{Data Statistics and Implementation Details}
\label{app:data}

Table~\ref{tab:data statistics} shows the detailed data statics of the training, development and test sets of the ACE05-E+ and ERE datasets. The statistics for the ERE dataset is slightly different from previous work~\cite{yinglinACL2020, text2event} as we consider the event triggers that are annotated with multiple types as different instances while the previous studies just keep one annotated type for each trigger span. For example, in the ERE-EN dataset, a token \textit{``succeeded''} in the sentence \textit{``Chun ruled until 1988 , when he was succeeded by Roh Tae - woo , his partner in the 1979 coup .''} triggers a \textit{End-Position} event of \textit{Chun} and a \textit{Start-Position} of \textit{Roh}. Previous classification based approaches did not predict multiple types for each candidate trigger.

%because we add back instances where multiple events triggered by the same span. For instances, in the ERE-EN dataset, a token \textit{``succeeded''} in the sentence \textit{``Chun ruled until 1988 , when he was succeeded by Roh Tae - woo , his partner in the 1979 coup .''} triggers a \textit{End-Position} event of Chun and a \textit{Start-Position} of Roh. Previous classification based approaches cannot deal with such situation.

\begin{table}[h]
\centering
\small
\begin{tabular}{lccc}
\toprule
Dataset &  Split & \# Events & \# Arguments
 \\
\midrule
\multirow{3}{*}{ACE05-E+}
    & Train &  4419  & 6605\\
    & Dev   &  468   & 757 \\
    & Test  &  424   & 689 \\
\midrule
% \multirow{3}{*}{ACE05-E}
%     & Train &  4202  & 4859\\
%     & Dev   &  450   & 605 \\
%     & Test  &  403   & 576 \\
% \midrule
\multirow{3}{*}{ERE-EN}
    & Train & 7394  & 11576 \\
    & Dev   & 632   & 979  \\
    & Test  & 669   & 1078  \\
\bottomrule
\end{tabular}
\caption{Data statistics for ACE2005 and ERE datasets.}
\label{tab:data statistics}
\end{table}

\paragraph{Zero-Shot Event Extraction} To evaluate the transfer capability of our approach, we use the top-$10$ most popular event types in ACE05 as seen types for training and treat the remaining 23 event types as unseen for testing, following~\citet{huang2017zero}. The top-$10$ training event types include \textit{Attack}, \textit{Transport}, \textit{Die}, \textit{Meet}, \textit{Sentence}, \textit{Arrest-Jail}, \textit{Transfer-Money}, \textit{Elect}, \textit{Transfer-Ownership}, \textit{End-Position}.  
We use the same data split as supervised event extraction but only keep the event annotations of the 10 seen types for training and development sets and sample 150 sentences with 120 annotated event mentions for the 23 unseen types from the test set for evaluation.

% \section{Implementation Details}
% \label{app:imp}

\paragraph{Implementation} For fair comparison with previous baseline approaches, we use the same pre-trained \texttt{bert-large-uncased} model for fine-tuning and optimize our model with BertAdam. We optimize the parameters with grid search: training epoch 10, learning rate $\in [3e\text{-}6, 1e\text{-}4]$, training batch size $\in\{8, 12, 16, 24, 32\}$, dropout rate $\in\{0.4, 0.5, 0.6\}$. Our experiments run on one Quadro RTX 8000. For trigger detection, the average runtime is 3.0 hours. For argument detection, the average runtime is 1.3 hours. We use Spacy to generate POS tags.
%\lifu{Add the details of entity detection model}

\begin{table*}[h]
\centering
\small
\begin{tabular}{lcc}
\toprule

    & {\textbf{Overlapped Triggers}} & {\textbf{Non-overlapped Triggers}}  \\ 
\midrule
OneIE~\cite{yinglinACL2020} & 88.2 & 71.0 \\
BERT\_QA\_Arg~\cite{xinyaduEMNLP2020} &  72.2 & 70.9 \\
\midrule
\textbf{Our Approach w/o Seed Triggers} & 88.9 & 70.8 \\
\textbf{Out Approach w/ Seed Triggers} & 97.2 & 71.3 \\
\bottomrule
\end{tabular}
\caption{Impact of seed triggers on supervised trigger extraction on ACE (F-score, \%)}
\label{tab:seed_impact}
\end{table*}
\begin{table*}[t!]
\centering
\small
{
\begin{tabular}{l|c|c|c|cc}
\toprule

\multirow{2}{*}{Source} & \multirow{2}{*}{Target} &\multirow{2}{*}{BERT\_QA\_Arg$_{\textrm{multi}}$ $\dagger$}&\multirow{2}{*}{BERT\_QA\_Arg$_{\textrm{binary}}$ $\dagger$}  &\multicolumn{2}{c}{\textbf{Our Approach} } \\ 

 &  & &  & w/o Seed Triggers & w/ Seed Triggers  \\ 
\midrule
ERE & ACE    & 48.9 & 50.8 & 53.8 & 53.9 \\
ACE & ACE    & 70.6 &72.2& 72.2 & 73.6 \\
ACE+ERE & ACE & 70.1 &71.3 & 72.2 & 74.4\\ \midrule
ACE & ERE    & 47.2 &47.2& 48.7 & 55.9 \\
ERE & ERE    & 57.0 & 56.7& 58.2 & 60.4\\
ACE+ERE & ERE  & 57.0 &54.6 & 56.2 & 63.0 \\
\bottomrule
\end{tabular}
}
\caption{Cross ontology transfer results for queries without seed triggers, between ACE and ERE datasets (F-score \%)}
%\lifu{In parenthesis, also add the performance on the ACE/ERE shared types on test datasets} \lifu{Reviewers may also ask what happen if we only use annotations for the overlapped types}
% \vspace{-4mm}
\label{tab:crossontology_seed}
\end{table*}

\paragraph{Evaluation Criteria} For evaluation of supervised event extraction, we use the same criteria as~\cite{qiliACl2013,chen2015event,nguyen_jrnn_2016,yinglinACL2020} as follows:
\begin{itemize}
    \item{\textbf{Trigger}: A trigger mention is correct if its span and event type matches a reference trigger.  Each candidate may act as triggers for multiple event occurrences. }
    
    \item{\textbf{Argument}: An argument prediction is correct only if the event trigger is correctly detected. Meanwhile, its span and argument role need to match a reference argument. An argument candidate  can be involved in multiple events as different roles. Furthermore, within a single event extent, an argument candidate may play multiple roles.}
    
    % \item{\textbf{Entity}: A Entity mention is correct if its span and entity type matches a reference trigger. }
\end{itemize}

\section{Impact of Seed Triggers}
\label{sec:impact_seed}
To investigate the impact of seed triggers on event trigger extraction, we take the supervised event extraction ACE dataset as a case study, where we divide the triggers in the evaluation dataset into two groups: overlapped triggers with the seeds or non-overlapped ones with the seeds. Then, we compare the performance of our approach with and without using seed triggers as part of the event type representations. As Table~\ref{tab:seed_impact} shows, by incorporating the seed triggers as part of the event type representations, our approach achieves better performance on both overlapped and non-overlapped triggers, demonstrating the benefit of seed triggers on representing event types. As the total number of overlapped triggers (34) is much lower than that of non-overlapped triggers (390), we view the impact of seed triggers on overlapped and non-overlapped triggers as comparable. On the other hand, by comparing our approach without using seed triggers with the BERT\_QA\_Arg baseline, our approach also achieves much better performance which is mostly due to the attention mechanism we used which can better capture the semantic consistency between the input tokens and the event type query which just consists of the event type name.

\section{In-depth Comparison for Cross Ontology Transfer}
\label{appendix:cross}
To deeply investigate the reason that our approach performs better than QA-based baselines from cross ontology transfer, we conducted ablation study by removing the seed triggers from the event type queries of our approach, as shown in Table~\ref{tab:crossontology_seed}. The BERT\_QA\_Arg$_{\textrm{multi}}$ utilizes a generic query, e.g., \textit{what's the trigger}, and classify each input token into one of the target types. It's essentially a multiclass classifier but just taking a query as the prompt. The BERT\_QA\_Arg$_{\textrm{binary}}$ utilizes each event type as the query to extract the corresponding event mentions. Comparing the two baseline methods, BERT\_QA\_Arg$_{\textrm{binary}}$ works slightly better than BERT\_QA\_Arg$_{\textrm{multi}}$, especially on ACE, demonstrating the benefit of type-oriented binary decoding mechanism. The only difference between BERT\_QA\_Arg$_{\textrm{binary}}$ and our approach without seed triggers is the learning of enriched contextual representations. The comparison of their scores demonstrates the effectiveness of the attention mechanisms designed for trigger extraction. Finally, by incorporating the seed triggers into event type representations, our approach is further improved significantly for all the settings. These in-depth comparisons demonstrate the effectiveness of both seed triggers and the attention mechanisms in our approach for transferring annotations from old types to the new types.

\section{More Ablation Studies of Supervised Event Extraction}
\label{sec:more_ablation}
The entity recognition model is based on a pre-trained BERT~\cite{devlin-etal-2019-bert} encoder with a CRF~\cite{lafferty2001conditional, passos-etal-2014-lexicon} based prediction network. It's trained on the same training dataset from ACE05 before event extraction, and the predictions are taken as input to argument extraction to indicate the candidate argument spans. Table~\ref{tab:Entity} shows the comparison of the entity extraction performance between our BERT-CRF approach and the baselines.

\begin{table}[h]
\centering
\small
\begin{tabular}{lr}
\toprule
    {\textbf{Model}} &  F1 \\ 
\midrule
    OneIE & 89.6\\
    FourIE & 91.1 \\
\midrule
  BERT+CRF  & 89.3 \\
\bottomrule
\end{tabular}
\caption{Performance of Entity Extraction (F-score, \%)}
\label{tab:Entity}
\end{table}

To understand the factors that affect argument extraction and decompose the errors propagated along the learning process (from predicted triggers or predicted entities), we conduct experiments that condition on given ground truth labels for those factors. Specifically, we investigate three settings: 1) given gold entity, 2) given gold event trigger, and 3) given both gold entity and event trigger. The experimental results is shown in Table \ref{tab:Partial}.

\begin{table}[h]
\centering
\small
\begin{tabular}{l|c|c}
\toprule
Given Information & ACE & ERE \\ 
\midrule
None & 55.1 & 50.2 \\
GE & 59.7 (+4.6) & 59.5 (+9.3) \\
GT & 68.7 (+13.6) & 67.2 (+17.0)\\
GT \& GE & 74.2 (+19.1) & 72.2 (+22.0)\\

\bottomrule
\end{tabular}
\caption{Performance of argument extraction conditioning on various input information: gold trigger (GT), and gold entities (GE). (F-score, \%)}
\label{tab:Partial}
\end{table}

\section{Remaining Challenges for Supervised Event Extraction}
\label{sec:remaining_challenges}
We sample $200$ supervised trigger detection and argument extraction errors from the ACE test dataset and identify the remaining challenges.

\paragraph{Lack of Background Knowledge} Background knowledge, as well as human commonsense knowledge, sometimes is essential to event extraction. For example, from the sentence ``\textit{since the intifada exploded in September 2000, the source said}'', without knowing that \textit{intifada} refers to a resistance movement, our approach failed to detect it as an \textit{Attack} event mention.

%Without external knowledge, our model is incapable of identifying proper nouns, such as 'intifada' in the following sentence:
%"\textit{since the intifada [GOLD: conflict attack] exploded in September 2000 , the source said}"
%. Suppose our model can learn from common sense knowledge graphs or pre-train on an external source, we can correct this error by recalling that 'intifada' refers to a resistance movement or rebellion.

\paragraph{Pronoun Resolution} Extracting arguments by resolving coreference between entities and pronouns is still challenging. For example, in the following sentence ``\textit{Attempts by Laleh and Ladan to have their operation elsewhere in the world were rejected, with doctors in Germany saying one or both of them could die}'', without pronoun resolution, our approach mistakenly extracted \textit{one}, \textit{both} and \textit{them} as \textit{Victims} of the \textit{Die} event triggered by \textit{die}, while the actual \textit{Victims} are \textit{Ladan} and \textit{Laleh}. 

%Distinguishing pronouns from actual arguments are difficult. For instance, in the following sentence "\textit{Attempts by Laleh and Ladan to have their operation elsewhere in the world were rejected , with doctors in Germany saying one or both of them could die}", our model mistakenly label "one", "both", and "them" as victims to the \textit{Life-Die} expressed by "die", while the actual victims are "Ladan" and "Laleh". 

\paragraph{Ambiguous Context} The ACE annotation guidelines~\cite{ldc_ace05} provide detailed rules and constraints for annotating events of all event types. For example, a \textit{Meet} event must be specified by the context as \textit{face-to-face and physically located somewhere}. Though we carefully designed several attention mechanisms, it is difficult for the machines to capture such context features accurately. For example, from the sentence ``\textit{The admission came during three-day talks in Beijing which concluded Friday, the first meeting between US and North Korean officials since the nuclear crisis erupted six months ago.}'', our approach failed to capture the context features that \textit{the talks is not an explicit face-to-face meet event}, and thus mistakenly identified it as a \textit{Meet} event mention.

\end{document}